\documentclass[10pt,twocolumn,letterpaper]{article}

\usepackage{iccv}              

\definecolor{iccvblue}{rgb}{0.21,0.49,0.74}
\usepackage[pagebackref,breaklinks,colorlinks,allcolors=iccvblue]{hyperref}
\usepackage[accsupp]{axessibility}  

\def\paperID{18} 
\def\confName{ICCV}
\def\confYear{2025}

\title{CoT-Pose: Chain-of-Thought Reasoning for 3D Pose Generation \\ from Abstract Prompts}

\author{Junuk Cha\thanks{Equal contribution.}\\
KAIST\\
South Korea\\
\and
Jihyeon Kim$^*$\\
KT\\
South Korea\\
}

\begin{document}
\maketitle
\begin{abstract}
Recent advances in multi-modal large language models (MLLMs) and chain-of-thought (CoT) reasoning have led to significant progress in image and text generation tasks. However, the field of 3D human pose generation still faces critical limitations. Most existing text-to-pose models rely heavily on detailed (low-level) prompts that explicitly describe joint configurations. In contrast, humans tend to communicate actions and intentions using abstract (high-level) language. This mismatch results in a practical challenge for deploying pose generation systems in real-world scenarios. To bridge this gap, we introduce a novel framework that incorporates CoT reasoning into the pose generation process, enabling the interpretation of abstract prompts into accurate 3D human poses. We further propose a data synthesis pipeline that automatically generates triplets of abstract prompts, detailed prompts, and corresponding 3D poses for training process. Experimental results demonstrate that our reasoning-enhanced model, CoT-Pose, can effectively generate plausible and semantically aligned poses from abstract textual inputs. This work highlights the importance of high-level understanding in pose generation and opens new directions for reasoning-enhanced approach for human pose generation.
\end{abstract}    
\section{Introduction}
\label{sec:intro}
Generating human poses from natural language descriptions has emerged as a key task at the intersection of human-centric vision and language, driven by its wide applicability in domains such as virtual character animation, human-robot interaction, and embodied AI. By enabling language-conditioned control over human motion, text-to-pose generation facilitates more intuitive and flexible human–machine interfaces. 

Recent efforts~\cite{briq2021towards, hong2022avatarclip, feng2024chatpose, li2025unipose, delmas2022posescript} have attempted to generate 3D human poses from text prompts. Some approaches~\cite{briq2021towards, hong2022avatarclip} rely on abstract prompts, such as ``generate the pose of throwing a ball" or ``generate the pose of playing the piano.", that depict the overall pose rather than providing detailed joint-level information. Although these abstract prompts are intuitive to humans, they are often ambiguous for deep learning networks. The lack of precise spatial cues makes it difficult to capture the exact pose, leading to poor generalization.

\begin{figure}[t]
    \centering
    \includegraphics[width=1\linewidth]{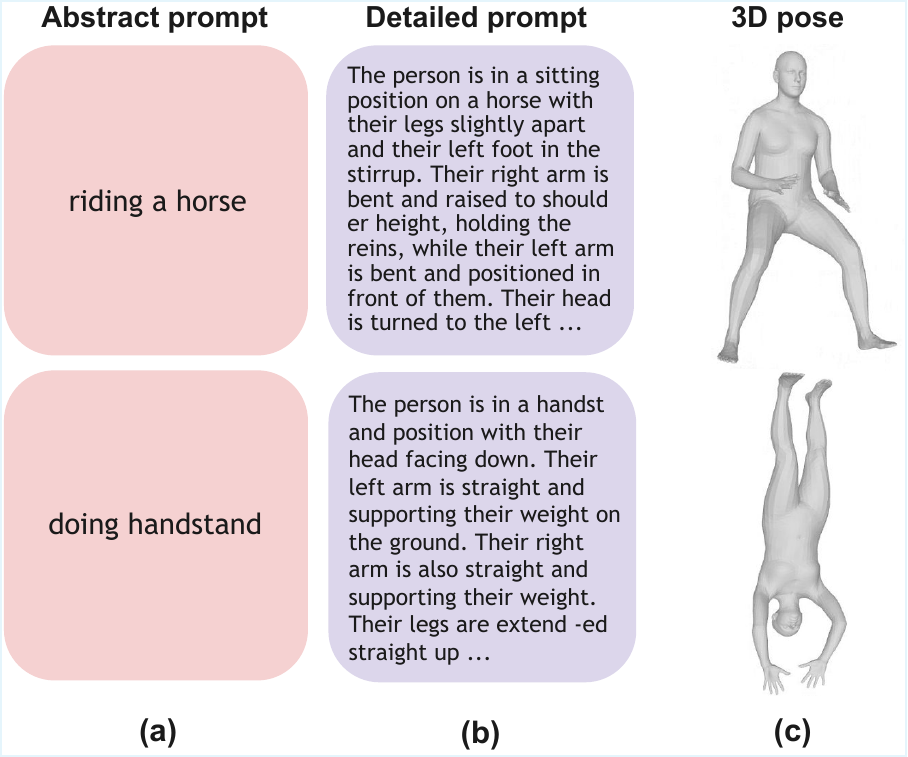}
    \caption{\textbf{From abstract to detailed prompts for 3D poses.} We illustrate (a) abstract prompts, (b) the corresponding detailed prompts, and (c) the ground-truth 3D pose mesh aligned with the abstract and detailed prompts. This highlights our objective of reasoning detailed prompts from abstract ones to enable accurate 3D human pose generation.}
    \label{fig:overview}
    \vspace{-3mm}
\end{figure}

Especially, AvatarCLIP~\cite{hong2022avatarclip} struggles to capture fine-grained semantics, such as distinguishing between left arm and right arm, highlighting its limited understanding of pose-relevant textual cues.
%
%
PoseScript~\cite{delmas2022posescript} introduced a language-annotated 3D pose dataset to facilitate semantic alignment between detailed textual descriptions and SMPL human poses~\cite{loper2023smpl}, and proposed a model that generates SMPL poses from such descriptions. However, in practical scenarios, humans generally rely on abstract prompts, which remains a limiting factor of PoseScript.

Both ChatPose~\cite{feng2024chatpose} and UniPose~\cite{li2025unipose} are capable of generating SMPL pose parameters from text prompts. While ChatPose generates continuous pose vectors, UniPose produces discrete pose tokens that are better aligned with the discrete text space. Benefiting from the unified pose-text representation, UniPose outperforms ChatPose on the text-to-pose generation task~\cite{li2025unipose}. However, since UniPose is trained with detailed textual descriptions from PoseScript~\cite{delmas2022posescript}, it requires such detailed prompts for pose generation, limiting its usability for general users who may prefer abstract prompts.


To generate 3D poses from abstract prompts, this paper leverages a chain-of-thought (CoT) reasoning technique that first generates detailed descriptions from abstract prompts and then sequentially produces 3D poses, inspired by reasoning-enhanced image generation~\cite{deng2025emerging}. To enable this capability, we fine-tune a multi-modal large language model (\eg, UniPose~\cite{li2025unipose}) to incorporate pose-specific reasoning capabilities. However, a key challenge is the lack of available training data for pose-specific reasoning. To address this, we propose a pipeline that automatically synthesizes triplets of abstract prompts, detailed prompts, and 3D SMPL poses~\cite{loper2023smpl}, with representative examples shown in Figure~\ref{fig:overview}. Specifically, we collect abstract prompts using ChatGPT~\cite{achiam2023gpt}, and utilize FLUX~\cite{labs2025flux} to generate images from the abstract prompts, SMPLest-X~\cite{yin2025smplest} to estimate 3D poses from the generated images, Pose2Text model~\cite{delmas2022posescript} to obtain detailed textual annotation from the estimated poses, and ChatGPT~\cite{achiam2023gpt} to refine detailed prompts. Finally, we fine-tune UniPose~\cite{li2025unipose} on our synthesized data to equip it with reasoning ability for pose generation.

Our contributions are summarized as follows:
\begin{itemize}
    \item We propose an automatic pipeline that synthesizes paired data consisting of abstract prompts, detailed prompts, and 3D SMPL poses.
    \item We propose a pose-specific reasoning model that employs chain-of-thought reasoning to generate intermediate detailed descriptions, which are then used to produce 3D SMPL poses from abstract prompts.
    \item We explore, for the first time to our knowledge, reasoning-enhanced 3D human pose generation from high-level abstract prompts.
\end{itemize}

\section{Related Work}

\noindent\textbf{Synthetic Datasets for Human Pose Generation.}
Creating large-scale datasets for the estimation or generation of 3D human poses is challenging due to the cost and difficulty of collecting accurate 3D annotations. To address this, several works have explored the use of synthetic data generated through simulation or rendering.
For instance, SURREAL~\cite{varol17_surreal} synthesizes SMPL-based human motions over real image backgrounds, providing pixel-level labels and 3D poses. 
AGORA~\cite{Agora2021} provides diverse human poses under realistic lighting and clothing variations.
However, their use in the text-to-pose research area is limited by the absence of paired textual descriptions.


%


\noindent\textbf{Text-to-Pose Generation.} Early approaches~\cite{delmas2022posescript,feng2024chatpose,li2025unipose} to generating 3D human poses from natural language have focused on using low-level pose descriptions as input. These descriptions specify detailed spatial configurations of body parts (e.g., ``the left arm is bent at 90 degrees"), allowing models to generate accurate SMPL~\cite{loper2023smpl} pose parameters. PoseScript~\cite{delmas2022posescript} introduced a large-scale dataset by automatically annotating SMPL meshes with such low-level text, and trained models to map between text and pose in both directions. UniPose~\cite{li2025unipose} further explored cross-modal representation learning by incorporating text, image, and pose information into a unified framework.

While effective in controlled settings, these methods rely on the availability of detailed pose descriptions at inference time, an assumption that is often impractical in real world applications. In everyday scenarios, users typically describe human posture using high-level, abstract expressions, such as ``a person is kicking" which lack the precise spatial details required for direct pose generation.

As a result, dependence on detailed textual inputs limits the scalability and usability of pose generation systems, particularly in open ended or user facing applications. This highlights the need for models that can interpret abstract prompts and internally reason about the underlying spatial structure to generate accurate SMPL pose parameters.

\noindent\textbf{Reasoning for Structured Generation.}
While chain-of-thought (CoT) prompting was originally introduced as a simple technique to induce step-by-step reasoning in large language models~\cite{wei2022chain, kojima2022large}, recent works have extended this idea to more complex forms of structured reasoning. Some methods maintain a prompt-based framework but incorporate compositional reasoning or planning to guide multi-step inference~\cite{zhou2022least}. Others advance further by embedding reasoning capabilities directly into the model architecture~\cite{jaech2024openai,guo2025deepseek}, employing a multi-modal reasoning pipeline~\cite{chen2024visual,deng2025emerging}, or an explicit cognitive structure~\cite{qi2025cogcom}.

These approaches demonstrate an advancement from simple prompting to reasoning-aware model architectures, reflecting the importance of intermediate reasoning in structured generation tasks. In the context of pose generation from abstract prompts, where implicit spatial reasoning is required to construct 3D structures, such modeling paradigms have the potential to support structured generation.



\begin{figure*}[t]
    \centering
    \includegraphics[width=1\linewidth]{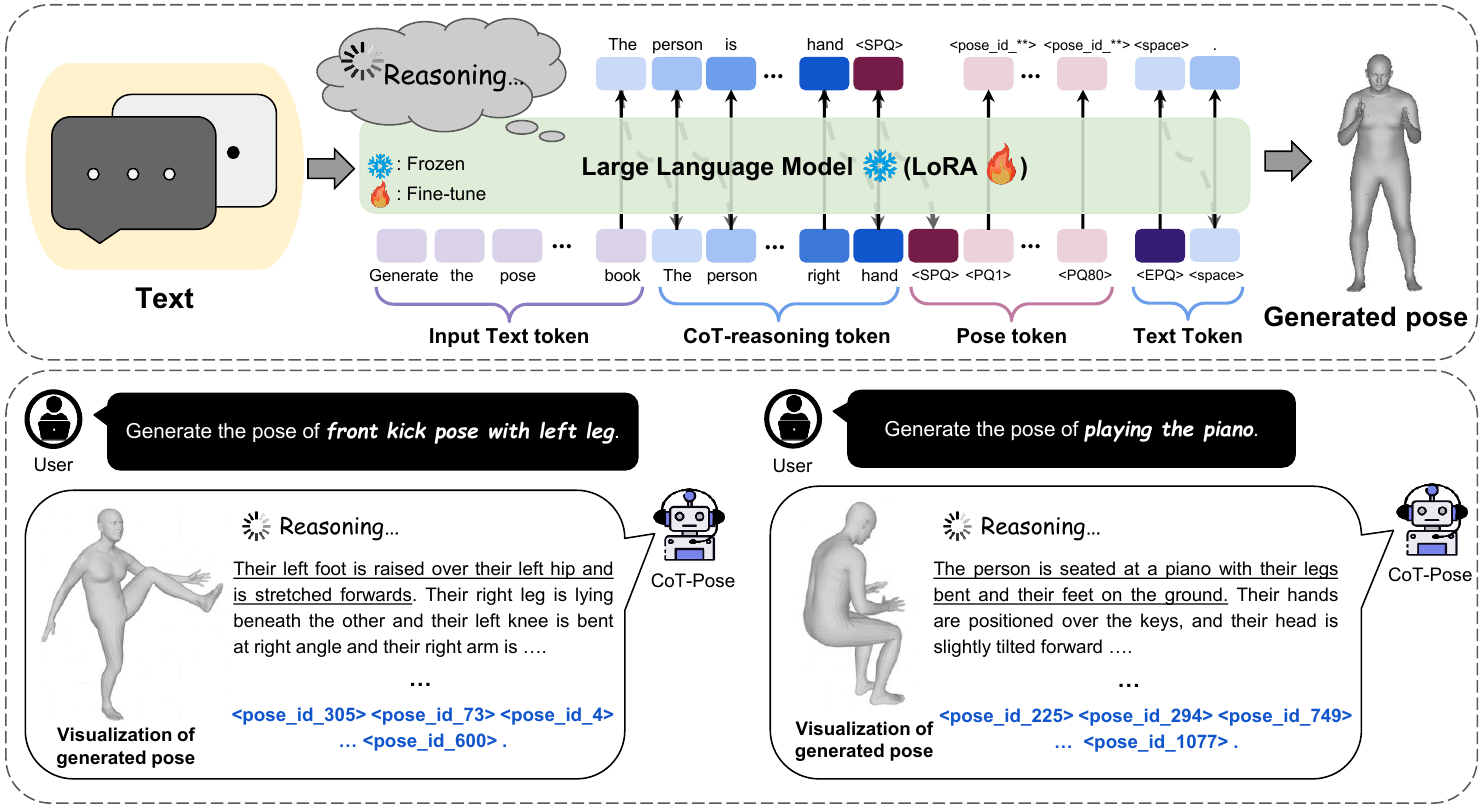}
    \caption{\textbf{Overview of the proposed CoT-pose framework and examples.} The top section presents the overall pipeline of CoT-Pose, which leverages large language models with chain-of-thought (CoT) reasoning to interpret abstract descriptions and generate semantically coherent 3D human poses. Given a prompt, the model first performs reasoning to decompose the abstract intent into detailed pose semantics, which are then used to generate corresponding 3D pose tokens. In the token sequence, SPQ and EPQ denote the start and end pose query tokens, respectively, while PQ represents an individual pose query token. The bottom part shows two inference examples where the model interprets abstract prompts via chain-of-thought reasoning to generate detailed descriptions, guiding the generation of semantically aligned 3D poses. The final poses are visualized on the left.}
    \label{fig:pipeline}
    \vspace{-3mm}
\end{figure*}

\section{Method}
Our method consists of two main components: an automatic data synthesis pipeline and a reasoning pipeline. Our automatic data synthesis pipeline integrates multiple models, including FLUX~\cite{labs2025flux}, SMPLest-X~\cite{yin2025smplest} and ChatGPT~\cite{achiam2023gpt}, to generate training data. The synthesized data is subsequently used to fine-tune UniPose~\cite{li2025unipose} with pose-specific reasoning capability. The overall inference process of our fine-tuned model is illustrated in Figure~\ref{fig:pipeline}.


\subsection{Automatic Data Synthesis}
\label{sec:automatic_data_synthesis}
In this pipeline, we utilize various models to generate the triplets consisting of abstract prompts, detailed prompts, and 3D SMPL~\cite{loper2023smpl} poses.

\noindent\textbf{Abstract Prompts.} We generate diverse action label descriptions that depict human action poses across a wide range of domains. To construct a comprehensive taxonomy, we first define the top-level categories of human actions by asking ChatGPT~\cite{achiam2023gpt} for a broad list of real-world activities. Based on its suggestions, we establish the following 11 top-level categories: \emph{Art and Performance}, \emph{Daily Activities}, \emph{Dance}, \emph{Fitness and Exercise}, \emph{Gestures and Communication}, \emph{Martial Arts}, \emph{Music}, \emph{Occupations and Professions}, \emph{Outdoor and Adventure}, \emph{Social Interactions}, and \emph{Sports}. For each top-level category, we further construct 50 sub-action classes to capture fine-grained variations in human poses. These sub-action classes are generated by leveraging ChatGPT with each category as context, ensuring diversity in motion types, action goals, and semantic interpretations. This results in a total of 550 unique action labels (\eg, ``checking mailbox", ``driving taxi", and ``running with baseball") spanning a broad spectrum of human activities. We transform each action label into an abstract prompt by prepending the phrase ``Generate the pose of”.

\noindent\textbf{3D SMPL Poses.} For text-to-pose generation, we need to collect 3D SMPL poses corresponding to the abstract prompts. While direct generation of 3D SMPL poses from abstract prompts is feasible, existing approaches~\cite{li2025unipose,delmas2022posescript,feng2024chatpose} struggle to produce high-quality results.

To address this issue, we utilize images as an intermediate representation. Specifically, we employ FLUX-LoRA-DLC~\cite{flux-lora-dlc} to generate realistic human images from abstract prompts. FLUX-LoRA-DLC is an open-source framework that integrates the FLUX model~\cite{labs2025flux} with LoRA-based fine-tuning~\cite{hu2022lora}, providing various LoRA variants adapted to diverse visual styles. Among them, we use the Super-Realism LoRA, which requires a special trigger word, ``Super Realism", to activate the corresponding visual domain during generation. In addition, we append the prompt, ``Full body, whole body. From head to feet.", to ensure that the generated images depict the entire human body from head to feet.

From the generated human images, we estimate 3D SMPL-X~\cite{pavlakos2019expressive} poses using SMPLest-X~\cite{yin2025smplest}. SMPLest-X is capable of estimating 3D poses even when some body parts are partially occluded or trimmed in the generated images. The estimated SMPL-X poses are then converted to SMPL~\cite{loper2023smpl} format to match the pose representation used by UniPose~\cite{li2025unipose}.

\noindent\textbf{Detailed Prompts.} From the estimated 3D SMPL poses, a pose-to-text captioning network~\cite{delmas2022posescript} generates detailed prompts that describe the configuration of each joint. However, due to orientation differences between the datasets used to train SMPLest-X~\cite{yin2025smplest} and the captioning network, the generated descriptions can sometimes be inaccurate. For example, a sitting pose may be misinterpreted as lying down.

To mitigate this issue, we utilize ChatGPT~\cite{achiam2023gpt} to refine the detailed prompts using both the action label and the generated image obtained from the abstract prompt as contextual cues.

\noindent\textbf{Filtering.} Based on the generated images, we manually filter out samples that lack meaningful human poses or contain inappropriate or sensitive content.


\subsection{Fine-tuning Multi-modal Large Language Model}
We fine-tune UniPose~\cite{li2025unipose} on our synthesized dataset to enable reasoning-enhanced text-to-pose generation. UniPose represents both text and pose in a shared discrete token space by discretizing continuous 3D poses into pose tokens through a trained pose tokenizer~\cite{dwivedi2024tokenhmr,van2017neural}. This unified embedding space enables joint modeling of language and pose within a single transformer framework. However, although the original UniPose framework supports text or pose generation depending on the task configuration, it does not allow joint generation of both modalities. In text-to-pose generation mode, only pose tokens are predicted, and the text serves solely as a conditioning input.

In contrast, our approach generates both the detailed description and the corresponding 3D pose within a single decoding pass from an abstract prompt. During reasoning, text tokens are decoded autoregressively with a causal attention mask, restricting each token to attend only to preceding tokens. After the reasoning phase, 80 pose query tokens are passed to the decoder input. The corresponding pose tokens are decoded in a single step using a bidirectional attention mask, enabling each token to attend to all other pose tokens and the entire preceding sequence of text tokens.

To enable joint generation of text and pose, we supervise the model using a combined objective function:
\begin{eqnarray}
    \mathcal{L} &=& \mathcal{L}_\text{text} + \mathcal{L}_\text{pose},\\
    \mathcal{L}_{\text{text}} &=& - \sum_{t=1}^{T} \log P\left(\mathbf{y}_t^{\text{detail}} \mid \mathbf{y}^{\text{abstract}}, \mathbf{y}_{<t}^{\text{detail}}; \theta \right), \\
\mathcal{L}_{\text{pose}} &=& - \log P\left(\mathbf{u} \mid \mathbf{y}^{\text{abstract}}, \mathbf{y}^{\text{detail}}, \mathbf{q}; \theta \right).
\end{eqnarray}

Here, $\mathcal{L}_{\text{text}}$ supervises the autoregressive generation of the detailed description $\mathbf{y}^{\text{detail}}$, conditioned on the abstract prompt $\mathbf{y}^{\text{abstract}}$ and the previously generated detailed text tokens $\mathbf{y}_{<t}^{\text{detail}}$. This guides the model to produce a sequence of semantically grounded instructions that describe fine-grained joint movements. The pose loss $\mathcal{L}_{\text{pose}}$ supervises the simultaneous prediction of all 80 pose tokens $\mathbf{u}$ in a single decoding step, conditioned on the abstract prompt $\mathbf{y}^{\text{abstract}}$, the generated detailed description $\mathbf{y}^{\text{detail}}$, and the learnable pose queries $\mathbf{q}$. This formulation allows the model to reason over the semantic content of the full description before decoding the spatial configuration of the human body.

By jointly optimizing both objectives, the model learns to align language and pose representations within a shared token space, while generating pose sequences that are semantically meaningful and structurally coherent.

\subsection{Decoding Pose Tokens to Mesh Vertices}
To reconstruct 3D human poses from the generated sequence of 80 discrete pose tokens, UniPose employs a VQ-VAE-based detokenizer~\cite{van2017neural,dwivedi2024tokenhmr}. Specifically, each pose token is first mapped to a corresponding codebook embedding. This embedding is then decoded into SMPL pose parameters of dimension $24 \times 3$, representing 24 joints in axis-angle format. These pose parameters are subsequently fed into the SMPL body model~\cite{loper2023smpl}, which outputs a posed 3D mesh consisting of 6,890 vertices and predefined face indices.

\section{Experiments}
\label{sec:experiments}

\subsection{Implementation Details}
We fine-tune UniPose~\cite{li2025unipose}, which is built on top of LLaVA-1.6V~\cite{liu2023visual}, using low-rank adaptation (LoRA)~\cite{hu2022lora}. LoRA is applied to selected modules within the transformer blocks, including the query, key, value, and output projections in the self-attention layers, as well as the up, down, and gate projections in the MLP layers. We set the LoRA rank at $r=64$, the scaling factor $\alpha=16$, and a dropout rate of 0.05, without applying any adaptation to bias terms. The model is trained for 5 epochs with a batch size of 8 and a learning rate of $5 \times 10^{-5}$.

\subsection{Dataset}
AMASS~\cite{mahmood2019amass} is a large-scale motion capture dataset that unifies diverse MoCap recordings from multiple datasets into a single representation using the SMPL body model, providing high-quality 3D human pose sequences without any linguistic annotations.
BABEL~\cite{punnakkal2021babel} extends AMASS by adding action-level natural language labels, which serve as abstract prompts, aligned with short motion clips, allowing coarse-grained semantic understanding of human activities.
PoseScript~\cite{delmas2022posescript} further bridges the gap between detailed prompts and 3D poses by densely annotating individual frames from AMASS with natural language descriptions. These annotations consist solely of low-level descriptions of the body configuration at each time step and do not include high-level abstract prompts.

We construct a dataset by temporally aligning abstract descriptions from BABEL and detailed descriptions from PoseScript with frame-wise SMPL annotations from AMASS, using frame indices as alignment references. In addition to the aligned dataset, we utilize our synthesized dataset, as described in Section~\ref{sec:automatic_data_synthesis}, to further fine-tune the model. We use 3,374 samples from the PoseScript dataset after removing duplicates based on action labels and 239 filtered samples from our synthesized dataset.

For evaluation, we collect a set of 50 real-world images from online sources, each paired with a corresponding abstract description used in the synthetic data generation. The 3D SMPL poses are estimated from these images using SMPLest-X~\cite{yin2025smplest}, and subsequently passed to the Pose2Text~\cite{delmas2022posescript} and ChatGPT~\cite{achiam2023gpt} to generate and refine detailed prompts, as described in Section~\ref{sec:automatic_data_synthesis}.

\subsection{Metrics}
We utilize pose feature distance (PFD) and mean per joint position error (MPJPE) to evaluate our method and compare it with state-of-the-art baselines, including PoseScript~\cite{delmas2022posescript}, ChatPose~\cite{feng2024chatpose}, and UniPose~\cite{li2025unipose}. 
PFD is computed as the distance between features extracted from ground-truth and generated poses using the pose encoder of the retrieval network~\cite{delmas2022posescript}.
MPJPE is measured as the average Euclidean distance between corresponding joints of the ground-truth and generated 3D poses.
For the ablation study, we additionally evaluate text feature distance (TFD) and multi-modality feature distance (MFD). Similar to PFD, TFD is computed as the distance between text features of ground-truth and generated low-level descriptions using the text encoder of the retrieval network~\cite{delmas2022posescript}. MFD measures the distance between the generated low-level descriptions and the corresponding generated poses, capturing the alignment across modalities.
Before computing all metrics, we align the global root rotation of the predicted pose with that of the ground-truth to remove differences in global orientation for all methods.
In our quantitative results (Tables~\ref{tab:quantitative_results}, \ref{tab:ablation_metrics}, and \ref{tab:ablation_metrics2}), we apply normalization factors to each metric for clearer presentation. Specifically, MPJPE is multiplied by 1000 to convert the unit from meters to millimeters, pose FID is scaled by 1000, and both text FID and multi-modality FID are scaled by a factor of 10. 

\begin{table}[t]
    \centering
    \begin{tabular}{l|cc}
    \hline
    Method & PFD ($\downarrow$) & MPJPE ($\downarrow$)\\
    \hline
    PoseScript~\cite{delmas2022posescript} & 1.0450 &  244.74 \\
    ChatPose~\cite{feng2024chatpose} & 0.6769 &  126.63 \\
    UniPose~\cite{li2025unipose} & 0.8760 &  172.89 \\
    Ours & \textbf{0.6162} & \textbf{124.91} \\
    \hline
    \end{tabular}
    \caption{Quantitative comparisons with the state-of-arts. Lower is better for both PFD and MPJPE, and MPJPE is measured in millimeters.}
    \label{tab:quantitative_results}
\end{table}


\begin{table}[t]
    \centering
    \small 
    \setlength{\tabcolsep}{7pt}
    \begin{tabular}{l|cccc}
    \hline
    Method & PFD ($\downarrow$) & TFD ($\downarrow$) & MFD ($\downarrow$) & MPJPE ($\downarrow$)\\
    \hline
    Ablation 1 & 0.7446 & \textbf{0.6268} & \textbf{0.6479} & 156.68 \\
    Ablation 2 & 0.6983 & 0.6355 & 0.6616 & 135.06 \\
    Ours       & \textbf{0.6162} & 0.6360 & 0.6639 & \textbf{124.91} \\
    \hline
    \end{tabular}
    \caption{Comparison of ablation settings and our full model across multiple metrics. Lower is better for all metrics, and MPJPE is measured in millimeters. Ablation 1 denotes training with only the reasoning text generation objective, while Ablation 2 denotes training without GPT-based detailed prompt refinement.}
    \label{tab:ablation_metrics}
\end{table}

\subsection{Quantitative Comparison}
We compare our approach with the existing methods, such as PoseScript~\cite{delmas2022posescript}, ChatPose~\cite{feng2024chatpose}, and UniPose~\cite{li2025unipose}, as shown in Table~\ref{tab:quantitative_results}. Our model achieves the best performance in metrics of pose feature distance (PFD) and mean per joint position error (MPJPE) across 50 abstract prompts.

PoseScript~\cite{delmas2022posescript} shows the worst performance, as it is trained solely on detailed prompts and struggles to interpret abstract prompts, resulting in poor reasoning of joint-level semantics. ChatPose~\cite{feng2024chatpose} achieves relatively good quantitative performance compared to other baselines, and its results are even comparable to ours. This is attributed to its generation of physically plausible, yet repetitive, and average-looking poses. UniPose~\cite{li2025unipose} fails to generate accurate poses due to its limited ability to interpret abstract prompts.

\begin{figure*}[t]
    \centering
    \includegraphics[width=0.88\linewidth]{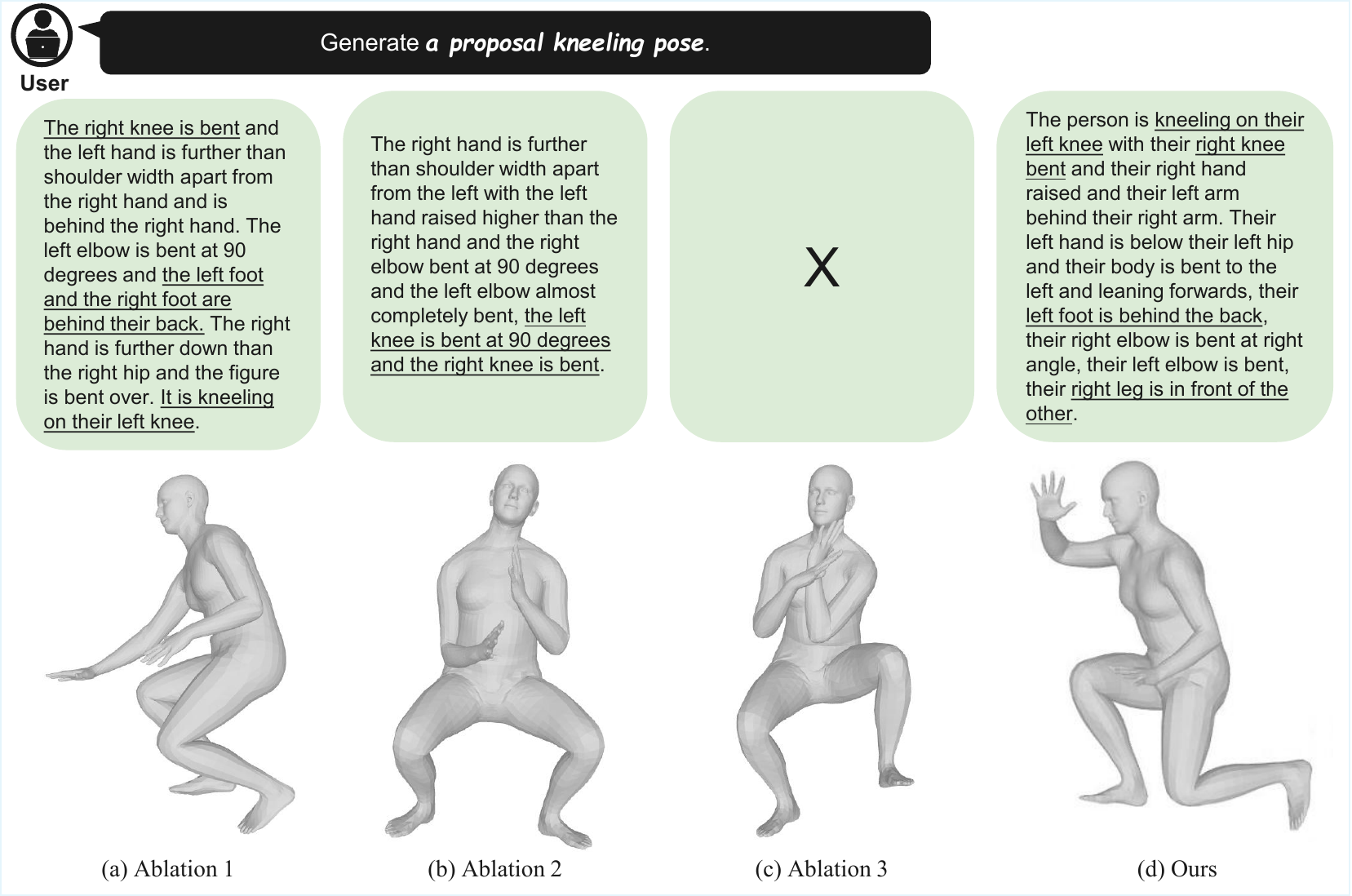}
    \vspace{-3mm}
    \caption{Qualitative comparison of the generation results for detailed prompts and 3D poses from (a) Ablation 1 (w/o pose supervision), (b) Ablation 2 (w/o prompt refinement), (c) Ablation 3 (w/o reasoning loss), and (d) Ours, conditioned on the same abstract prompt. Our method produces the most accurate and coherent outputs, demonstrating the effectiveness of reasoning and prompt refinement.}
    \label{fig:qualitative_ablation}
    \vspace{-3mm}
\end{figure*}

\subsection{Qualitative Comparison}
\begin{figure*}
    \centering
    \includegraphics[width=0.88\linewidth]{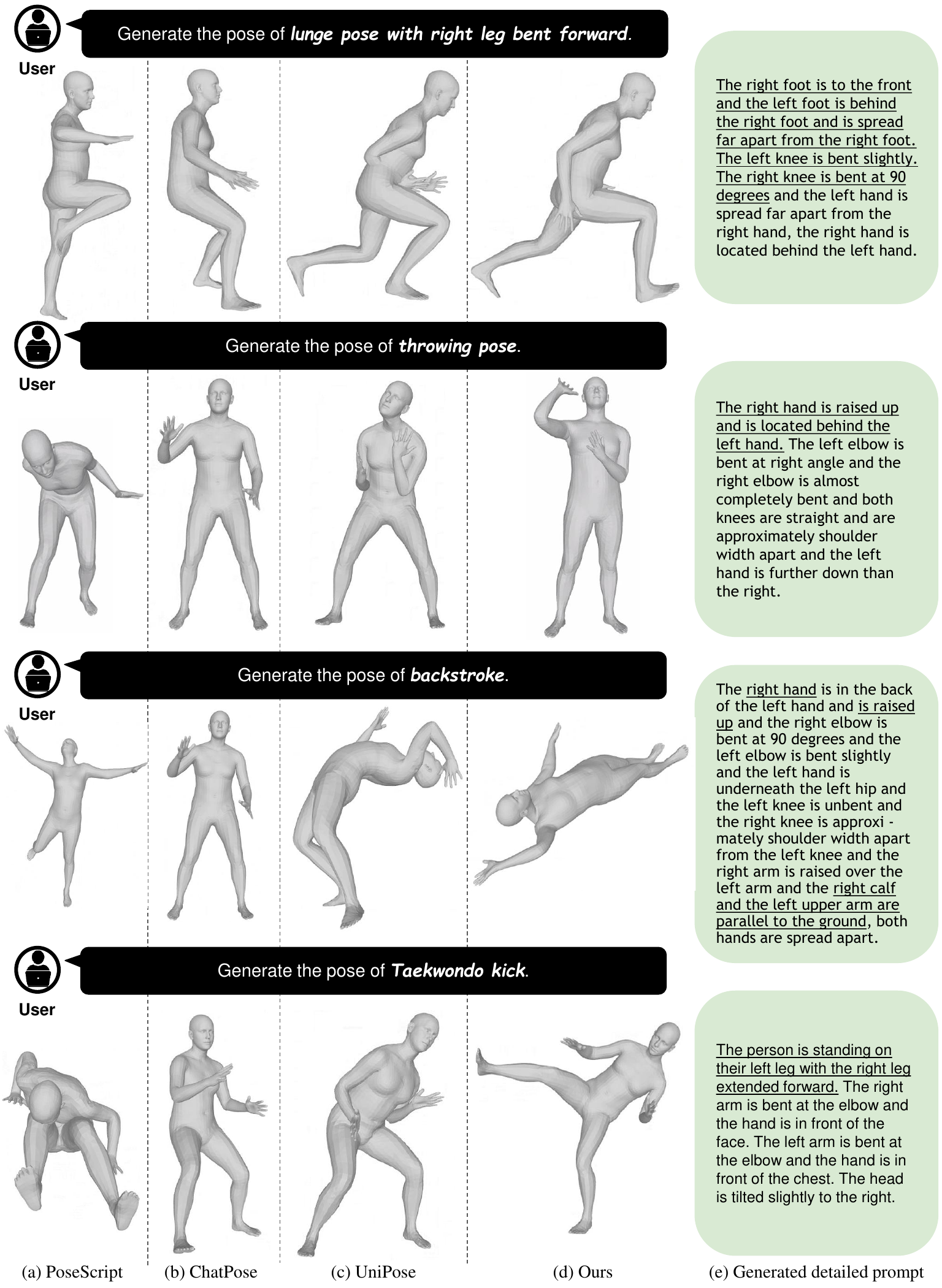}
    \vspace{-2mm}
    \caption{Qualitative comparison of 3D pose generation results from (a) PoseScript~\cite{delmas2022posescript}, (b) ChatPose~\cite{feng2024chatpose}, (c) UniPose~\cite{li2025unipose}, and (d) Ours, given the same abstract prompt. (e) shows the detailed prompt generated by our method from the abstract prompt. Prior methods (a–c), which directly condition on abstract prompts, often produce implausible poses with unnatural bending or occlusions, indicating a limited understanding of high-level intent. In contrast, our method (d) first infers a detailed prompt (e) and then generates a 3D pose by reasoning about the joint-level motion required to realize the abstract prompts.}
    \label{fig:qualitative}
\end{figure*}

We present a qualitative comparison in Figure~\ref{fig:qualitative}, demonstrating the results of representative text-to-pose generation methods, including PoseScript\cite{delmas2022posescript}, ChatPose~\cite{feng2024chatpose}, UniPose~\cite{li2025unipose}, and our proposed approach.

PoseScript~\cite{delmas2022posescript}, trained exclusively on detailed prompts, fails to interpret abstract pose instructions and produces the most degraded outputs among the four methods, often exhibiting severe self-occlusions and anatomically implausible bending. ChatPose~\cite{feng2024chatpose} avoids anatomically implausible artifacts; however, it does not capture semantic distinctions between abstract prompts, resulting in highly similar poses in rows 2–4. This suggests a limited understanding of the semantics of abstract instructions or a lack of diversity in generation. UniPose~\cite{li2025unipose} attempts to capture variations in abstract prompts more effectively than the preceding methods, yet still suffers from generation performance and artifacts such as unnatural joint bending and partial occlusions.

In contrast, our method explicitly reasons over the abstract prompt by first generating an intermediate detailed prompt, which enables accurate synthesis of the corresponding 3D pose. As shown in Figure~\ref{fig:qualitative} (e), the detailed prompt bridges the semantic gap between high-level human instructions and low-level pose representations. This comparison highlights the importance of modeling the correspondence between user-friendly abstract prompts and machine-interpretable pose descriptions and demonstrates the effectiveness of our approach in closing this gap.

\begin{table}[t]
    \centering
    \small 
    \begin{tabular}{l|cc}
    \hline
    Method & PFD ($\downarrow$) & MPJPE ($\downarrow$)\\
    \hline
    Ablation 3 & \textbf{0.5701} & \textbf{115.14} \\
    Ours       & 0.6162 & 124.91 \\
    \hline
    \end{tabular}
    \caption{Comparison of ablation setting and our full model. Lower is better for all metrics, and MPJPE is measured in millimeters. Ablation 3 denotes training with only the pose token generation objective.}
    \label{tab:ablation_metrics2}
    \vspace{-3mm}
\end{table}

\begin{figure*}[t]
    \centering
    \includegraphics[width=1\linewidth]{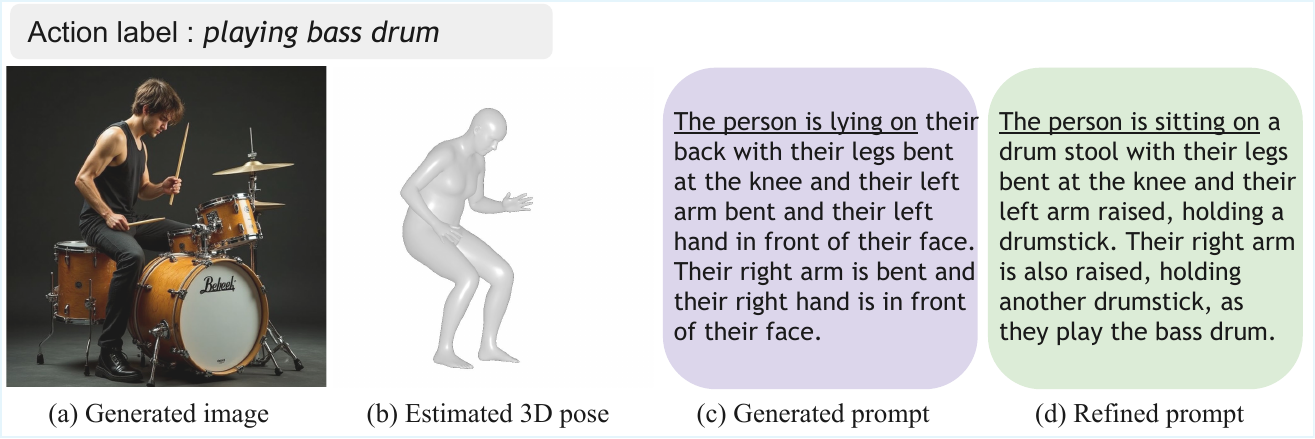}
    \vspace{-7mm}
    \caption{Illustration of data synthesis process for the action playing bass drum, using our synthesis pipeline based on FLUX~\cite{flux-lora-dlc}, SMPLest-X~\cite{yin2025smplest}, Pose2Text~\cite{delmas2022posescript}, and ChatGPT~\cite{achiam2023gpt}. (a) shows the generated image, and (b) presents the estimated 3D pose. (c) displays the initially generated prompt, which inaccurately describes the pose. (d) shows the refined prompt that better aligns with the action label, 3D pose, and generated image.}
    \label{fig:dataset}
    \vspace{-3mm}
\end{figure*}

\subsection{Ablation Study}
\subsubsection{Design of the Data Synthesis Module} 
We automatically construct training triplets consisting of an abstract prompt, a detailed prompt, and its corresponding 3D SMPL pose using our proposed pipeline.

Generating 3D SMPL poses directly from abstract prompts using existing methods, such as Text2Pose~\cite{delmas2022posescript}, ChatPose~\cite{feng2024chatpose}, and UniPose~\cite{li2025unipose}, remains challenging, as shown in Figure~\ref{fig:qualitative}. This limitation motivates our pipeline, which introduces images as an intermediate representation to facilitate the generation of semantically consistent 3D SMPL poses.

We present the data synthesis process in Figure~\ref{fig:dataset}. The process begins by generating an image from the abstract prompt using FLUX~\cite{flux-lora-dlc}. The 3D pose is then extracted using SMPLest-X~\cite{yin2025smplest}, and the corresponding detailed textual description is generated using Pose2Text~\cite{delmas2022posescript}. However, the raw outputs from Pose2Text often fail to accurately reflect the physical plausibility of the pose, particularly under abstract prompts. To address this, we introduce a refinement step using ChatGPT~\cite{achiam2023gpt}, which improves the prompt quality and ensures better alignment across detailed prompt, 3D pose, and action label. This results in well-structured and semantically coherent training triplets.

\subsubsection{Fine-tuning Strategy}
We conduct experiments to evaluate the effectiveness of jointly learning detailed prompt generation and pose generation (Ablation 1: without pose token generation loss $\mathcal{L}_\text{pose}$), and the impact of ChatGPT-based prompt refinement (Ablation 2), as demonstrated in Table~\ref{tab:ablation_metrics}.
In Ablation 1, we fine-tune the model using only the text token generation loss $\mathcal{L}_{\text{text}}$, while still providing the pose generation trigger token (the start pose query token, denoted as SPQ in Figure~\ref{fig:pipeline}) during inference, which is the same token used in the full model to prompt pose token generation.
In Ablation 2, we fine-tune the model using the detailed prompts generated by the Pose2Text model~\cite{delmas2022posescript}, which exhibit misalignment in global orientation.




Furthermore, we investigate the effectiveness of the reasoning step through an additional experiment, as shown in Table~\ref{tab:ablation_metrics2} using our collected real-image dataset with prompts seen during training, and in Figure~\ref{fig:qualitative_ablation} using unseen abstract prompts. Remarkably, even without fine-tuning the text token generation loss $\mathcal{L}_{\text{text}}$ (Ablation 3), which implies the absence of explicit reasoning supervision, the model achieves competitive quantitative results. However, as illustrated in Figure~\ref{fig:qualitative_ablation}, it generates semantically inconsistent or implausible poses when evaluated on novel abstract prompts unseen during training. This discrepancy highlights the limited generalization capability of the model without chain-of-thought reasoning, underscoring the necessity of explicit reasoning supervision for robust abstract prompt-to-pose generation.

In addition, as shown in Figure~\ref{fig:qualitative_ablation}, Ablation 1 focuses solely on generating detailed prompts from abstract inputs, without any supervision for pose generation. Consequently, it fails to produce valid 3D poses. Ablation 2 replaces the refined detailed prompts generated by ChatGPT~\cite{achiam2023gpt} with the unrefined output of the Pose2Text model~\cite{delmas2022posescript}. Without well-structured intermediate guidance, this variant frequently generates semantically inconsistent and less accurate poses. Overall, these results confirm the importance of both explicit reasoning supervision and prompt refinement in achieving robust and accurate 3D pose generation from abstract prompts.


\section{Conclusion}
With the advancement of multi-modal large language models (MLLMs) and chain-of-thought (CoT) reasoning, numerous fields have experienced significant progress. 
In this work, we first propose a method for automatically synthesizing triplet data consisting of abstract prompts, detailed prompts, and 3D human poses. This dataset enables structured supervision for pose generation.
Building on our dataset and prior work, we introduce CoT-Pose, a reasoning-enhanced framework that infers detailed prompts from abstract ones and subsequently produces accurate 3D human poses. While our approach demonstrates promising results, future work includes scaling up the training data to develop stronger foundation models, improving the generation fidelity of fine-grained hand poses, and addressing orientation inconsistencies.

\vspace{3mm}
\noindent\textbf{Acknowledgements.}
J. Cha was supported by Culture, Sports and Tourism R\&D Program through the Korea Creative Content Agency grant funded by the Ministry of Culture, Sports and Tourism in 2024 (Project Name: Development of barrier-free experiential XR contents technology to improve accessibility to online activities for the physically disabled, Project Number: RS-2024-00396700).
{
    \small
    \bibliographystyle{ieeenat_fullname}
    \bibliography{main}
}

\end{document}